# Ambiguous Proximity Distribution

Quanquan Wang[1,2*] , Yongping Li[1]

[1]Shanghai Institute of Applied Physics, Chinese Academy of Sciences, Shanghai, 201800, China
[2] University of Chinese Academy of Sciences, Bejing 100049, China
{wangquanquan,ypli}@sinap.ac.cn

**Abstract.** Proximity Distribution Kernel is an effective method for bag-of-featues based image representation. In this paper, we investigate the soft assignment of visual words to image features for proximity distribution. Visual word contribution function is proposed to model ambiguous proximity distributions. Three ambiguous proximity distributions is developed by three ambiguous contribution functions. The experiments are conducted on both classification and retrieval of medical image data sets. The results show that the performance of the proposed methods, Proximity Distribution Kernel (PDK), is better or comparable to the state-of-the-art bag-of-features based image representation methods.

## 1   Introduction

Due to the rapid growth of modern digital imaging and video technologies, the volume of visual data is also increasing significantly. The management and retrieval of visual information has attracted much attention of the research community of computer vision, machine learning, and scientific computing [1-10]. Recently, Content-Based Image Retrieval (CBIR) has been one of the most popular research areas [11–14, 63, 64]. Given an query image, this problem is to retrieval images from a large image database based on the visual similarity between the query and the database images. In this case, representing the visual information of each images as some feature measures are critical [15–20]. Most often, both global and local visual information are important. Cognitive science community has shed some light on this topic, in particular in the ground laying work of Kong [21, 22] on how human visual cognition processes global and local visual information and utilize both type of information to solve complex visual problems. Kong's work on global/local visual information processing has not only made great impact in the field of cognitive science, but also inspired many advances outside cognitive science, such as in the one discussed in this paper [21, 22]. J. Yang et al. [23] presented a novel approach for foreground objects removal while ensuring structure coherence and texture consistency. The approach used structure as a guidance to complete the remaining scene. The work benefits a wide range of applications especially for the online massive collections of imagery, such as photo localization and scene reconstructions.

Moreover, this work is applied to privacy protection by removing people from the scene. Inspired by Kong's work [21, 22], various image representation methods have been proposed. Among them, local visual descriptors has been shown to be outperforming other visual methods [24–28]. This kind of methods are also called bag-of-features since each images is represented as a collection of local features [29–31].

This work presents a novel content-based image retrieval system, which is based on the Visual Words (VW) framework. VW framework is a recently introduced framework, and it has been successfully applied to scenery and object image retrieval tasks [32, 33, 15]. The VW model represents each image using a discrete and disjunct visual vocabulary which is composed of many visual word elements [34-37]. Base on this model, it is possible to split an image into a set of visual features and to represent the image using the statistics of these local visual features. The ideal intrinsic features should be transition, rotation, and scaling invariant [38, 39]. However, in many applications, it is hard to find the ideal features. In our system the visual features are the key points with SIFT (Scale Invariant Feature Transform) descriptors [40–43], or image patches (small sub images) [44–47]. One effective statistical image presentation of VM model is Proximity Distribution (PD) [48, 49]. Moreover, a corresponding kernel function is also proposed to match a pair of proximity distributions. The PD is defined as the distributions of co-occurring local features as they appear at increasing distances from one another. One inherent procedure of the PD model is the discretization of visual words from continuous image local features.

In this study we investigate the effect of ambiguity modeling on ambiguous proximity model. To the best of our knowledge, there are no published studies on the application of visual word ambiguity in constructing proximity distribution. The main contribution of this paper is an investigation of visual word ambiguity leading to explicit *Visual Word Contribution Function* for the *Ambiguous Proximity Distribution* model. Moreover, we try to develop a kernel to match two Ambiguous Proximity Distribution, so that it can be used in image retrieval.

The paper is organized as follows: In section 2, we introduce the proposed Ambiguous Proximity Distribution Kernel. Section 2.1 discusses the original Proximity Distribution Kernel model. Section 2.2 introduces the Visual Word Contribution Function for generalization of hard and soft assignment of a local feature to a word in the dictionary, and by developing three ambiguous contribution functions, we proposed three *Ambiguous Proximity Distributions* in section 2.3. Section 3 presents the experimental results and Section 4 concludes with the final remarks.

## 2 Ambiguous Proximity Distribution Kernel

Let $X$ be an image and $\{x_l\}$; $l = 1, \cdots, L$ a collection of $L$ local features extracted from $X$. Typically, the local features or regions $x_l$ are detected interest regions with SIFT descriptors [40], or densely sampled image patches [50]. In the learning phase, we construct a codebook $V$ using a clustering algorithm. Usually, k-means is used to cluster centers of features which extracted from all images in the database, then, these cluster centers are used as a vocabulary (codebook) $V = \{v_1, \cdots, v_K\}$ with $K$ visual words for all images to get word vector representation [51–53]. On the set of all available features of query image $X$, we perform vector-quantization, to arrive at $K$ codebook elements $V = \{v1, \cdots, vK\}$. So, in this first coding stage, the image $X$ is represented by the local feature $\{(xl, \alpha l)\}k = 1, \cdots, L$, with each $\_l$ identified with the integers $i = 1, \cdots, K$.

$$\alpha_l = \arg\min_i(D(vi, xl)) \tag{1}$$

where $xl$ is image region $l$, and $D(vi, xl)$ is the distance between a codeword $vi$ and region $xl$.

Given a vocabulary of $K$ of codewords, and the traditional VW approach describes an image $X$ by a distribution over the visual words. For each word $vi$ in the vocabulary $V$, the traditional VW model estimates the distribution of visual words in an image by

$$H_{CB}^{X}(i) = \frac{1}{L}\sum_{l=1}^{L} I(\alpha_l = i) \tag{2}$$

The indicator $I(x)$ outputs 1 when the Boolean variable $x$ is true and 0 otherwise, so that,

$$I(\alpha_l = i) = \begin{cases} 1, & \text{if } vl = \arg\min_{vj \in V} D(vj, xl) \\ 0, & \text{otherwise} \end{cases} \tag{3}$$

By applying $I(\alpha_l = i)$, the region $xl$ is only assigned to the nearest word $vi$ in the dictionary. The VW model represents an image by a histogram of word frequencies that describes the probability density over codewords.

### 2.1  Proximity Distribution Kernel

The *proximity distribution kernel*(PDK) is proposed by Ling and Soatto in [54], which matches distributions of co-occurring local features as they appear at increasing distances from one another in an image $X$. In this case, each feature point $xl$ is first mapped to one of $K$ discrete visual words $vi \in V$, which are the prototypical local features identified via clustering on local features of many training images.

Each local feature set $\{xl, \alpha_l\}$, $l = 1, \cdots, L$ of $X$ is converted to a $K \times K \times R$-dimensional histogram. This histogram counts the number of times each visual word co-occurs within the $r = 1, \cdots, R$ spatially nearest neighbor features of any other visual word. Specifically, for a given image input represented as a local feature set $X$, each histogram element $H^X(I, j, r)$ is defined as the number of times visual word type $j$ occurs within the $r$ spatially nearest neighbors of a visual word of type $i$. The proximity distribution of $X$ is given as

$$H_{PD}^X(i,j,r) = \#\{(\alpha_l, \alpha_m): \alpha_l = i, \alpha_m = j, d_{NN}(x_l, x_m) \leq r\}$$
$$= \sum_{l=1}^{n}\sum_{m=1}^{n} I(\alpha_l = i)I(\alpha_m = j)I(d_{NN}(x_l, x_m) \leq r) \quad (4)$$

where $d_{NN}(x_l, x_m) \leq r$ indicates that $x_m$ is within the $r$-th nearest neighbors of $x_l$. R is the size of the neighborhood. Since $r = 1, \cdots, R$, apparently it is a cumulative distribution of the cooccurring pairs of words. The (unnormalized) proximity distribution kernel (PDK) value between two images with feature sets $Y$ and $Z$ is then

$$K_{PDK}(Y, Z) = \sum_{i=1}^{K}\sum_{j=1}^{K}\sum_{r=1}^{R} \min(H^Y(i,j,r), H^Z(i,j,r)) \quad (5)$$

where $H_Y$ and $H_Z$ are the associated arrays of histograms computed for the two input feature sets for image $Y$ and $Z$.

## 2.2 Visual Word Contribution Function

In this section, we generalize the bag-of-features based image presentation as a visual word contribution form. First of all, for a region $x_l$ in an image $X$, we define it's contribution in the constructing of an statistical presentation (histogram or proximity distribution) as *Visual Word Contribution Function*. In detail, in the vector quantization phase, the region $x_l$ will only be assigned to the nearest visual word $v_i$ in the dictionary $V$ as $\alpha_i = \arg\min_k(D(v_k, x_i))$ in (2), while not considering any other visual words $\{v_j | (D(v_j, x_i)) > D(v_k, x_i); j \neq k\}$. When constructing the traditional VW based histogram $H_{CB}$ in the accumulative way in (2), the quantities of $x_l$'s contribution in accumulating the $i$-th bin of the histogram $H_{CB}(i)$ is $I(\alpha_l = i)$ as (3). Here, we define the *Visual Word Contribution Function* of $x_l$ for $v_i$ as its contribution in accumulating the $i$-th bin as $F(v_i, x_l)$, so that for coodbook base histogram can be rewritten as

$$H_{CB}^X(i) = \frac{1}{L}\sum_{l=1}^{L} F(v_i, x_l), \text{ and } F(v_i, x_l) = I(\alpha_l = i) \quad (6)$$

Then we consider the Visual Word Contribution for Proximity Distribution $H_{PD}$. Apparently, by $F(v_i, x_l) = I(\alpha_l =$, (4) can be rewritten as

$$H_{PD}^X(i,j,r) = \sum_{l=1}^{n}\sum_{m=1}^{n} F(v_i, x_l)F(v_j, x_m)I(d_{NN}(x_l, x_m) \leq r) \quad (7)$$

In the above histogram and proximity distribution, one inherent component of the *hard assignment* for codebook model is the assignment of image local feature to visual words in the vocabulary by using *Hard Assignment Contribution Function* as $F_{hard}(v_i, x_l)$

$$F_{hard}(v_l, x_l) = \begin{cases} 1, & \text{if } \alpha_{l=l} \\ 0, & \text{otherwise} \end{cases} \quad (8)$$

Here, an important assumption is that a discrete visual word is a characteristics representative of a continuous image feature.

**2.3    Ambiguous Proximity Distribution**

The continuous nature of local visual features indicates selecting a most representative visual word from a visual vocabulary. It is possible that one local feature can have zero, one, or multiple optimal visual words in a visual vocabulary [34]. If there is only one optimal visual word, there is no ambiguity. We propose a robust alternative method of discrete proximity distribution to estimate a probability density function, which is based on kernel density estimation [34]. We call it *Ambiguous Proximity Distribution*. By giving three different ambiguous contribution function to replace the hard contribution function $F_{hard}$, we developed three kinds of Ambiguous Proximity
Distributions (APD) as follows:

**Kernel Proximity Distribution** In the VW model, the histogram estimation function (presented by hard contribution function Fhard) of the visual words may be replaced by a kernel density estimation function. Moreover, a suitable kernel (such as the Gaussian kernel $K_\sigma(x) = \frac{1}{\sqrt{2\pi}\sigma} \exp(-\frac{1}{2}\frac{x^2}{\sigma^2})$ ,where σ is the bandwidth parameter of kernel $K_\sigma(x)$ ) is used, and it allows kernel density estimation to become a part of the visual word. With this function, the *Kernel Proximity Distribution* is given as,

$$H_{ker}^X(i,j,r) = \sum_{l=1}^{L}\sum_{m=1}^{L} \{F_{ker}(v_i, x_l) F_{ker}(v_j, x_m) I(d_{NN}(x_l, x_m) \le r)\}$$
$$= \sum_{l=1}^{L}\sum_{m=1}^{L} \{K_\sigma(D(v_i, x_l)) K_\sigma(D(v_j, x_m)) I(d_{NN}(x_l, x_m) \le r)\} \quad (9)$$

where $F_{ker}(v_i, x_l)$ is the *Kernel Contribution Function*, as follows,

$$F_{ker}(v_i, x_l) = K_\sigma(D(v_i, x_l)) = \frac{1}{\sqrt{2\pi}\sigma} \exp(-\frac{1}{2}\frac{D(v_i, x_l)^2}{\sigma^2}) \quad (10)$$

In this way, a kernel visual vocabulary replaces the hard mapping of local features in an image region to the visual vocabulary. This soft assignment models is realized by replacing the hard contribution function with the kernel contribution function. This soft assignment models two types of ambiguity between visual words: *codeword* *uncertainty* and *codeword plausibility*, which are introduced as follows.

**Uncertainty Proximity Distribution** Uncertainty Contribution Function indicates that one image local feature $x_l$ may be assigned to more than one visual words. *Uncertainty Contribution Function* is defined as follows,

$$F_{unc}(v_i, x_l) = \frac{K_\sigma(D(v_i, x_l))}{\sum_{j=1}^{|V|} K_\sigma(D(v_j, x_l))} \tag{11}$$

$F_{unc}$ normalizes the probabilities to 1 and is distributed over all visual words. By using $F_{unc}$, we can define the *Uncertainty Proximity Distribution* as

$$\begin{aligned} H_{unc}^X(i,j,r) &= \sum_{l=1}^{L} \sum_{m=1}^{L} \{K_{unc}(w_i, x_l) K_{unc}(w_j, x_m) I(d_{NN}(x_l, x_m) \leq r)\} \\ &= \sum_{l=1}^{L} \sum_{m=1}^{L} \{\frac{K_\sigma(D(v_i, x_l))}{\sum_{k=1}^{|V|} K_\sigma(D(v_k, x_l))} \frac{K_\sigma(D(v_j, x_l))}{\sum_{k=1}^{|V|} K_\sigma(D(v_k, x_l))} I(d_{NN}(x_l, x_m) \leq r)\} \end{aligned} \tag{12}$$

In this way, visual word uncertainty is kept and it has the ability to assign one local feature to multiple visual words. However, it does not take the plausibility of a visual word into account.

**Plausibility Proximity Distribution** Visual word Plausibility Contribution Function is proposed by by using ambiguous contribution functions. This function indicates that an image local feature may not be close enough to be represented by all the visual words in the vocabulary. *Plausibility Contribution Function* is defined as

$$F_{pla}(v_i, x_l) = K_\sigma(D(v_i, x_l)) I(\alpha_l = l) = \begin{cases} K_\sigma(D(v_i, x_l)), & \text{if } \alpha_l = i \\ 0, & \text{otherwise} \end{cases} \tag{13}$$

$F_{pla}$ selects for an image local feature $x_l$ only the closet visual word $v_i$ and assigns it probability to the kernel value of that visual word. But it should be noted that it cannot select multiple visual word candidates.

In this way, we redefine the Proximity Distribution using Plausibility Contribution Function $F_{pla}$, resulting the *Plausibility Proximity Distribution* as

$$\begin{aligned} H_{pla}^X(i,j,r) &= \sum_{l=1}^{L} \sum_{m=1}^{L} \{F_{pla}(v_i, x_l) F_{pla}(v_j, x_m) I(d_{NN}(x_l, x_m) \leq r)\} \\ &= \sum_{l=1}^{L} \sum_{m=1}^{L} \{K_\sigma(D(v_i, x_l)) I(\alpha_l = i) K_\sigma(D(v_j, x_m)) I(\alpha_m = j) I(d_{NN}(x_l, x_m) \leq r)\} \end{aligned} \tag{14}$$

An unified formula is given as follows for these three ambiguous proximity distributions:

$$H^X(i,j,r) = \sum_{l=1}^{L}\sum_{m=1}^{L}\{F(w_i, x_l)F_{pla}(w_j, x_m)I(d_{NN}(x_l, x_m) \leq r)\} \quad (15)$$

where $F(v_i, x_l)$ is to the a version of $K_{ker}(v_i, x_l)$ for Kernel Proximity Distribution, $F_{unc}(v_i, x_l)$ for Uncertainty Proximity Distribution, or $F_{pla}(v_i, x_l)$ for Plausibility Proximity Distribution respectively. Moreover, by setting $F(v_i, x_l) = F_{hard}(v_i, x_l) = I(\alpha_l = i)$, the original hard assignment of Proximity Distribution in (4) can also be included in (15).

Then the corresponding (unnormalized) *Ambiguous Proximity Distribution Kernel* (APDK) value between two images with feature sets $Y$ and $Z$ is

$$K_{apd}(Y, Z) = \sum_{i=1}^{K}\sum_{j=1}^{K}\sum_{r=1}^{R}\min(H_{apd}^Y(i,j,r), H_{apd}^Z(i,j,r)) \quad (16)$$

where $H_{apd}^Y$ and $H_{apd}^Z$ are the associated arrays of *Ambiguous Proximity Distribution* computed for the two input feature sets.

## 3    Experiments

The proposed method is in general applicable for CBIR of large image databases according to learned local visual vocabulary. In this section, we demonstrate our algorithm using image patch exemplars [55] applied to a nearest neighbor classification problem medical image retrieval. We carry the experiments based on ImageClef 2007 medical image classification and ImageClef 2008 large-scale medical image retrieval competitions. In these two groups of experiments, each image is treated as a collection of image local image features.

### 3.1    Experiment on ImageClef 2007 Dataset

In the ImageClef 2007 medical image classification competition [56], a database of 12,000 categorized radiograph images is used. In this experiment, we also use this database. A set of 11,000 images are used as training set, and the remaining 1000 images are used as test images. There are 116 different classes within this database, based on the differences of either the examined region, the image orientation with respect to the body or the biological system under evaluation.

We represent each image as a bag of small patches, which are the representation of local features of an image. We extract a small patch around every pixel, using a patch size of 9 × 9 pixels. and the level of noise, we have also applied the Principal Component Analysis (PCA) [57, 58] procedure to the data dimensionality of descriptor vector of each local feature from 81 to 7. The next step of our system is to learn a vocabulary of visual words based on a set of local features of images. The main step in the vocabulary construction procedure is clustering the patches, and we use the the k-means algorithm. A small-size vocabulary of visual words are generated. Using the generated visual vocabulary, each image $X$ is represented as an Ambiguous Proximity

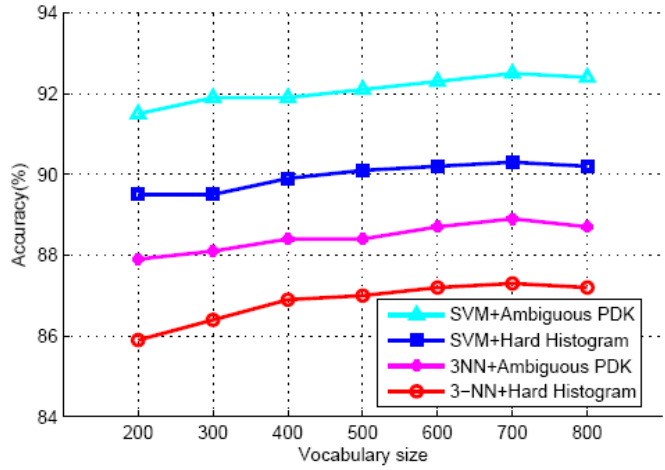

(a)

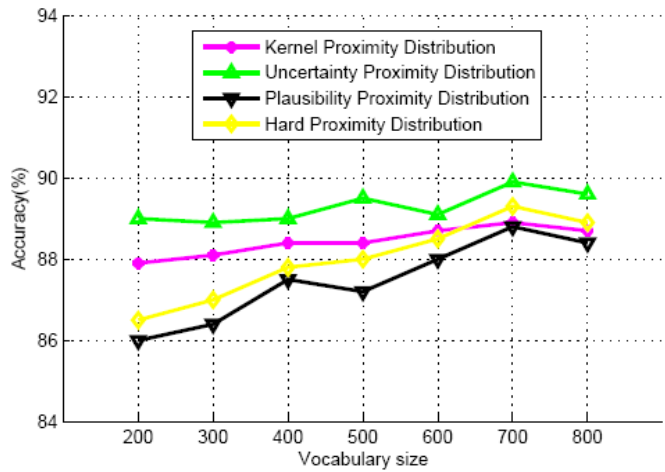

(b)

Fig. 1 (a) Effect of vocabulary size, for K-NN and SVM classifiers. (b) Classification performance results of various types of Ambiguous PDK for the ImageClef 2007 dataset over various vocabulary sizes

Distribution ( $H^X_{KER}$, $H^X_{UNC}$ or $H^X_j$ ), a Hard Proximity Distribution $H$, or a histogram of visual words. In this step images are sampled with a dense grid. We firstly use the *k* nearest neighbor classifier for the classification problem [59–61]. Given a query image the retrieval is based on finding the nearest image in the labeled training set. With our Ambiguous Proximity Distribution, we adopt the Ambiguous Proximity Distribution Kernel $K_{APD}$ to compute the similarity of the Ambiguous Proximity Distribution of two patches collections for images. We use the $L_1$ norm distance $D^{L1}$ ($H^X$, $H^Y$) between the word histograms of the two images *X* and *Y* as distance measure for hard histogram,

$$D^{L1}(H^X, H^Y) = \sum_{k=1}^{K} | h_k^X - h_k^Y | \qquad (17)$$

We have found that we gain much better results using a multi-class SVM classifier. The multiclass SVM is implemented as a series of one-vs-one binary SVMs with a Ambiguous Proximity Distribution Kernel $K_{APD}$. We run 20 cross-validation experiments trained on 10000 training images and then test it on 1000 randomly selected test images.

As Figure 1 shows, we increase the number of visual words up to 700 words. The performance of vocabulary size of 700 visual words is significantly better than that of the 200 vocabulary for the Hard Histogram baseline. However, with Ambiguous PDK (Ambiguous Proximity Distribution Kernel) the results are very similar for both sizes, as Ambiguous PDK can utilize more information of provided by vocabulary by Ambiguous Visual Word Contribution Function. Figure 1 (a) also demonstrates that using an SVM classifier provides results that are more than 3% higher than the best K-NN classifier for both Hard Histogram baseline and Ambiguous PDK. We then make a concrete analysis of the results in Fig. 1 (b) with the various types of Ambiguous Proximity Distributions. The results show that Uncertainty Proximity Distributions $H_{UNC}$ consistently outperforms other types of ambiguity for all vocabulary sizes. But we should note that this performance gain is not always that significant. Moreover, for a vocabulary size of 200, Uncertainty Proximity Distributions $H_{UNC}$ outperforms Hard Assignment based Proximity Distributions. On the other end of the performance scale, there is Plausibility Proximity Distribution $H_{PLA}$, which always yields the lowest results. We can also see that a Kernel Proximity Distribution $H_{KER}$, outperforms Hard Proximity Distributions for smaller vocabulary sizes.

### 3.2 Large Scale Image Retrieval Experiment

In ImageClef 2008 a large-scale medical image retrieval competition was conducted. In this competition, there are 66,000 images in the database, and also 30 query topics. Moreover, each topic is composed of one or more images and also a short textual description. The objective of retrieval is to return a set of 1000 images from the given database. Figure 2 shows the scores of our Ambiguous Proximity Distribution Kernel (Ambiguous PDK), along with visual retrieval algorithms submitted by additional groups [62]. As the results shown in Fig. 2, The Ambiguous PDK for either original method shows the best possible performance, and again as we decrease the returned images, we can expect more accurate results at the cost of fewer returned images.

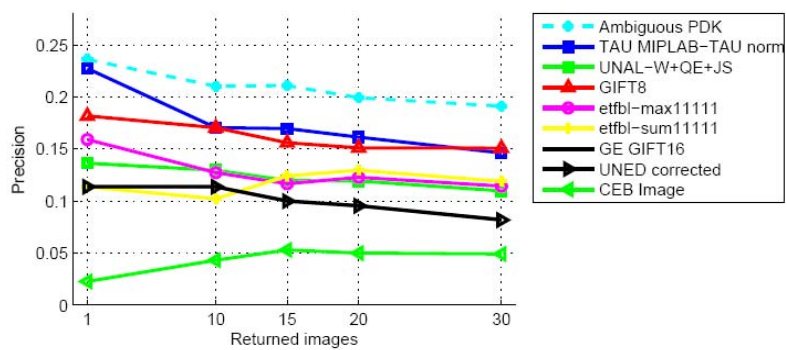

Fig. 2 Precision vs Recall graph of visual retrieval systems; ImageClef 2008 medical database. Predication shown for first 5,10,15,20 and 30 returned images.

## 4    Conclusions

This paper proposed Ambiguous Proximity Distribution, a novel image presentation combining the advantage of visual word ambiguity and ambiguous proximity distribution. A visual word contribution function framework was used for analyzing its relation with the popular VW model and developing three novel Ambiguous Proximity Distributions. An extensive comparative medical image retrieval and classification experimental analysis with state-of-the art medical image retrieval methods provided empirical evidence of the effectiveness of the proposed technique for enhancing the performance of medical image retrieval system.